\documentclass[runningheads,a4paper]{llncs}
\newcommand{\bsym}{\boldsymbol}
\newcommand{\mbf}{\mathbf}

\usepackage{graphicx}
\usepackage{amsmath,amssymb} 
\usepackage{booktabs}
\usepackage{multirow}
\usepackage{subfigure}
\usepackage{pdfcomment}

\renewcommand{\Re}{\mathbb{R}}

\begin{document}
\mainmatter  

\title{A Unified Framework for Probabilistic Component Analysis}

\titlerunning{A Unified Framework for Probabilistic Component Analysis}

\author{Mihalis A. Nicolaou$^1$
\and Stefanos Zafeiriou$^1$ \and Maja Pantic$^{1,2}$}
\authorrunning{A Unified Framework for Probabilistic Component Analysis}

\institute{$^1$Department of Computing, Imperial College London, UK\\
$^2$EEMCS, University of Twente, NL\\
{\tt \{mihalis, s.zafeiriou, m.pantic\}@imperial.ac.uk\\}
}

\maketitle
\begin{abstract}
\looseness-1We present a unifying framework which reduces the construction of probabilistic component analysis techniques to a mere selection of the latent neighbourhood, thus providing an elegant and principled framework for creating novel component analysis models as well as constructing probabilistic equivalents of deterministic component analysis methods.  Under our framework, we unify many very popular and well-studied component analysis algorithms, such as Principal Component Analysis (PCA), Linear Discriminant Analysis (LDA), Locality Preserving Projections (LPP) and Slow Feature Analysis (SFA), some of which have no probabilistic equivalents in literature thus far.  We firstly define the Markov Random Fields (MRFs) which encapsulate the latent connectivity of the aforementioned component analysis techniques; subsequently, we show that the projection directions produced by all PCA, LDA, LPP and SFA are also produced by the Maximum Likelihood (ML) solution of a single joint probability density function, composed by selecting one of the defined MRF priors while utilising a simple observation model.  Furthermore, we propose novel Expectation Maximization (EM) algorithms, exploiting the proposed joint PDF, while we generalize the proposed methodologies to arbitrary connectivities via parametrizable MRF products.   Theoretical analysis and experiments on both simulated and real world data show the usefulness of the proposed framework, by deriving methods which well outperform state-of-the-art equivalents.
\keywords{Unifying Framework, Probabilistic Methods, Component Analysis, Dimensionality Reduction, Random Fields}
\end{abstract}
\section{Introduction}
\looseness-1Unification frameworks in machine learning provide valuable
material towards the deeper understanding of various methodologies, while also they form a flexible basis upon which further extensions can be easily built. One of the first attempts to unify methodologies was made
in \cite{RoweisZoubin}.
In this seminal work, models such as Factor Analysis (FA), Principal
Component Analysis (PCA), mixtures of Gaussian clusters, Linear Dynamic Systems, Hidden Markov Models
and Independent Component Analysis were unified as variations of
unsupervised learning under a single basic generative model.

\looseness-1Deterministic Component Analysis (CA) unification frameworks proposed in previous works, such as \cite{akisato2013designing}, \cite{kokiopoulou2011trace}, \cite{Borga92aunified}, \cite{de2012least} and \cite{sun2009least}, provide significant insights on how CA methods such as Principal Component Analysis, Linear Discriminant Analysis, Laplacian Eigenmaps and others can be jointly formulated as, e.g., least squares problems under mild conditions or general trace optimisation problems.  Nevertheless, while several probabilistic equivalents of, e.g. PCA have been formulated (c.f., \cite{TippingPPCA} \cite{roweis1998algorithms}), to this date no unification framework has been proposed for {\it probabilistic} component analysis.  \looseness-1Motivated by the latter, in this paper we propose the {\it first} unified framework for probabilistic component analysis.  Based on Markov Random Fields (MRFs), our framework unifies {\it all} component analysis techniques whose corresponding deterministic problem is solved as a trace optimisation problem without domain constraints for the parameters, such as Principal Component Analysis (PCA), Linear Discriminant Analysis (LDA), Locality Preserving Projections (LPP) and Slow Feature Analysis (SFA).  Our framework provides further insight on  component analysis methods from a probabilistic perspective.  This entails providing probabilistic explanations for the data at hand with explicit variance modelling, as well as reduced complexity compared to the deterministic equivalents.  These features are especially useful in case of methods for which no probabilistic equivalent exists in literature so far, such as LPP.  Furthermore, under our framework one can generate {\it novel} component analysis techniques by merely combining products of MRFs with arbitrary connectivity. 

\looseness-1The rest of this paper is organised as follows.  We initially introduce previous work on CA, highlighting the properties of the proposed framework (Sec. \ref{sec:prior_art}).  Subsequently, we formulate the joint complete-data Probability Density Function (PDF) of observations and latent variables. We show that the Maximum Likelihood (ML) solution of this joint PDF is co-directional to the solutions obtained via deterministic PCA, LDA, LPP and SFA, by changing only the prior latent distribution (Sec. \ref{sec:ML}), which, as we show, models the latent dependencies and thus determines the resulting CA technique.  E.g, when using a fully connected MRF, we obtain PCA. When choosing the product of a fully connected MRF and an MRF connected only to within-class data, we derive LDA. LPP is derived by choosing a locally connected MRF, while finally, SFA is produced when the joint prior is a linear Markov-chain.  Based on the aforementioned PDF we subsequently propose Expectation Maximization (EM) algorithms (Sec. \ref{sec:aunifiedEMforCA}). Finally in Sec. \ref{sec:experiments}, utilising both synthetic and real data, we demonstrate the usefulness and advantages of this family of probabilistic component analysis methods.
\section{Prior Art and Novelties}
\label{sec:prior_art}
\looseness-1An important contribution of our paper lies in the proposed unification
of probabilistic component techniques, giving  rise to the first framework that reduces the construction of probabilistic component analysis models to the design of a appropriate prior, thus defining only the latent neighbourhood.  Nevertheless, other novelties arise in methods generated via our framework.
In this section, we review the
state-of-the-art in deterministic and probabilistic PCA, LDA, LPP and SFA.
While doing so, we highlight novelties and advantages that our
proposed framework entails wrt. each alternative formulation.
Throughout this paper we consider,  a
zero mean set of $F$-dimensional observations of length $T$, represented by the matrix
$\mathbf{X} = [\mathbf{x}_1,\ldots,\mathbf{x}_T]$. All CA methods
discover an $N$-dimensional latent space $\mathbf{Y} =
[\mathbf{y}_1,\ldots,\mathbf{y}_T]$ which preserves certain properties
of $\mathbf{X}$.

\subsection{Principal Component Analysis (PCA)}
The deterministic model of PCA finds a set of projection bases
$\mathbf{W}$, with the latent space $\mathbf{Y}$ being the projection of
the training set $\mathbf{X}$ (i.e., $\mathbf{Y} =
\mathbf{W}^T\mathbf{X}$)). The optimization problem is as follows
\begin{equation}
\mathbf{W}_o = \arg\max_{\mathbf{W}} \mbox{tr}\left[ \mathbf{W}^T
\mathbf{S} \mathbf{W} \right], \,\, \mbox{s.t.}\,\,
\mathbf{W}^T\mathbf{W} = \mathbf{I}
\end{equation}
where $\mathbf{S} = \frac{1}{T} \sum_{i=1}^T \mathbf{x}_i\mathbf{x}_i^T$
is the total scatter matrix and $\mbf{I}$ the identity matrix. The optimal $N$ projection basis
$\mathbf{W}_o$ are recovered (the $N$ eigenvectors of $\mathbf{S}$ that
correspond to the $N$ largest eigenvalues).
\looseness-1Probabilistic PCA (PPCA) approaches were independently proposed in
\cite{roweis1998algorithms} and \cite{TippingPPCA}. In
\cite{TippingPPCA} a probabilistic generative model was adopted as:
\begin{equation}
\begin{array}{rl}
\mathbf{x}_i = \mathbf{W}\mathbf{y}_i + \mbox{\boldmath$\epsilon$}_i,
\,\, \mathbf{y}_i \sim \mathcal{N}(\mathbf{0},\mathbf{I}), \,\,
\mbox{\boldmath$\epsilon$}_i \sim \mathcal{N}(\mathbf{0},\sigma^2\mathbf{I})
\end{array}
\end{equation}
\looseness-1 where $\mathbf{W} \in \Re^{F \times N}$ is the matrix that relates the
latent variable $\mathbf{y}_i$ with the observed samples $\mathbf{x}_i$
and $\mbox{\boldmath$\epsilon$}_i$ is the noise
which is assumed to be an isotropic Gaussian model. The motivation is
that, when $N < F$, the latent variables will offer a more parsimonious
explanation of the dependencies arising in observations.

\subsection{Linear Discriminant Analysis (LDA)} 

Let us now further assume that our data $\mathbf{X}$ is further
separated into $K$ disjoint classes $\mathcal{C}_1,\ldots,\mathcal{C}_K$
 with $T = \sum_{c=1}^K |\mathcal{C}_c|$. The Fisher's Linear Discriminant Analysis
(LDA) finds a set of projection bases $\mathbf{W}$ s.t. \cite{yan2007graph}
\begin{equation}\label{eq:ldaopt}
\begin{array}{rl}
\mathbf{W}_o & = \arg\min_{\mathbf{W}} \mbox{tr}\left[ \mathbf{W}^T
\mathbf{S}_w \mathbf{W} \right], \,\, \mbox{s.t.}\,\,
\mathbf{W}^T\mathbf{S}\mathbf{W} = \mathbf{I}
\end{array}
\end{equation}
where $\mathbf{S}_w=\sum_{c=1}^K\sum_{\mbf{x}_i\in
\mathcal{C}_c}(\mathbf{x}_i-\boldsymbol\mu_{\mathcal{C}_i})(\mathbf{x}_i-\boldsymbol\mu_{\mathcal{C}_i})^T$ and $\boldsymbol\mu_{\mathcal{C}_i}$ the mean of class $i$.
The aim is to find the latent space $\mathbf{Y}=\mathbf{W}^T\mathbf{X}$
such that the within-class variance is minimized in a whitened space. The
solution is given by the eigenvectors of $\mathbf{S}_w$ corresponding to the $N-K$ smallest eigenvalues of the whitened data. \footnote{We adopt this
formulation of LDA instead of the equivalent of
maximizing the trace of the between-class scatter matrix
\cite{belhumeur1997eigenfaces}, since this facilitates our following
discussion on Probabilistic LDA alternatives.}

\looseness-1Several probabilistic latent variable models which exploit class
information have been recently proposed (c.f.,
\cite{PLDAidentity,Zhang2009,IoffeECCV2006}). In
\cite{PLDAidentity,Zhang2009} another two related attempts were made to
formulate a PLDA. Considering $\mathbf{x}_i$ to be the $i$-th sample of the
$c$-th class, the generative model of \cite{PLDAidentity} can be
described as:
\begin{equation}
\mathbf{x}_i  = \mathbf{F}\mathbf{h}_c + \mathbf{G}\mathbf{w}_{ic} +
\mbox{\boldmath$\epsilon$}_{ic}, \,\, \mathbf{h}_c,\mathbf{w}_{ic} \sim
\mathcal{N}(\mathbf{0},\mathbf{I}), \,\, \mbox{\boldmath$\epsilon$}_{ic} \sim
\mathcal{N}(\mathbf{0},\mathbf{\Sigma})
\end{equation}
where $\mathbf{h}_c$ represents the class-specific weights  and
$\mathbf{w}_{ic}$ the weights of each individual sample, with $\mbf{G}$ and $\mbf{F}$ denoting the corresponding loadings. Regarding
\cite{Zhang2009}, the probabilistic model is as follows:
\begin{equation}\label{E:PLDA1}
\mathbf{x}_i = \mathbf{F}_c\mathbf{h}_c +
\mbox{\boldmath$\epsilon$}_{ic}, \,\, \mathbf{h}_c,\mathbf{F}_{ic} \sim
\mathcal{N}(\mathbf{0},\mathbf{I}), \,\, \mbox{\boldmath$\epsilon$}_{ic} \sim
\mathcal{N}(\mathbf{0},\mathbf{\Sigma})
\end{equation}
We note that the two models become equivalent when choosing a common
$\mathbf{F}$ (Eq. \ref{E:PLDA1}) for all classes while also disregarding
the matrix $\mathbf{G}$. In this case, the ML solution is given by
obtaining the eigenvectors corresponding to the largest eigenvalues of $\mathbf{S}_w$. Hence, the
solution is vastly different than the one obtained by deterministic LDA
(which keeps the smallest ones, Eq. \ref{eq:ldaopt}), resembling more to the solution of problems which retain the maximum variance. In fact, when learning a different $\mathbf{F}_c$
per class, the model of \cite{Zhang2009} reduces to applying PPCA per class.
To the best of our knowledge the only probabilistic model where the ML
solution is closely related to that of deterministic LDA is
\cite{IoffeECCV2006}. The probabilistic model is defined as follows:
$\mathbf{x} \in \mathcal{C}_i$, $\mathbf{x}|\mathbf{y} \sim
\mathcal{N}(\mathbf{y},\mathbf{\Phi}_w)$, $\mathbf{y} \sim
\mathcal{N}(\mathbf{m},\mathbf{\Phi}_b)$, $\mathbf{V}^T\mathbf{\Phi}_b\mathbf{V} =
\mathbf{\Psi}$ and $\mathbf{V}^T\mathbf{\Phi}_w\mathbf{V} = \mathbf{I}$,
$\mathbf{A} = \mathbf{V}^{-T}$, $\mathbf{\Phi}_w =
\mathbf{A}\mathbf{A}^T$ $\mathbf{\Phi}=\mathbf{A}\mathbf{\Psi}\mathbf{A}^T$,
where the observations are generated as:
\begin{equation}\label{E:PLDA2}
\begin{array}{r}
\mathbf{x}_i= \mathbf{A}\mathbf{u}, \,\, \mathbf{u} \sim
\mathcal{N}(\mathbf{V},\mathbf{I}), \,\, \mathbf{v} \sim \mathcal{N}(\mathbf{0},\mathbf{\Psi}).
\end{array}
\end{equation}
The drawback of \cite{IoffeECCV2006} is the requirement for all classes to contain
the same number of samples.  As we show, we overcome this limitation in our formulation.

\subsection{Locality Preserving Projections (LPP)}\label{S:LPP}

\looseness-1Locality Preserving Projections (LPP) is the linear alternative to
Laplacian Eigenmaps \cite{niyogi2004locality}. The aim is to obtain
a set of projections $\mathbf{W}$ and a latent space $\mathbf{Y} =
\mathbf{W}^T\mathbf{X}$ which preserves the locality of the
original samples. First, let us define a set of weights that represent
locality. Common choices for the weights are the heat kernel $u_{ij} =
e^{-\frac{||\mathbf{x}_i - \mathbf{x}_j||^2}{\gamma}}$ or a set of constant
weights ($u_{ij} = 1$ if the $i$-th and the $j$-th vectors are adjacent
and $u_{ij} = 0$ otherwise, while $u_{ij}=u_{ji}$).  LPP finds a set of projection basis matrix
$\mathbf{W}$ by solving the following problem:
\begin{equation}
\begin{array}{rl}
\mathbf{W}_o & = \arg\min_{\mathbf{W}} \sum_{i,j=1}^T
\sum_{n=1}^N u_{ij}|| \mathbf{w}_n^T\mathbf{x}_i -
\mathbf{w}_n^T\mathbf{x}_j||^2 \\
& = \arg\min_{\mathbf{W}} \mbox{tr}\left[ \mathbf{W}^T \mathbf{X}
\mathbf{L} \mathbf{X}^T \mathbf{W} \right] \mbox{s.t.} \,\, \mathbf{W}^T\mathbf{X}\mathbf{D}\mathbf{X}^T\mathbf{W}
= \mathbf{I}
\end{array}
\end{equation}
where $\mbf{U}=[u_{ij}]$, $\mathbf{L} = \mathbf{D} - \mathbf{U}$ and $\mathbf{D} =
\mbox{diag}(\mathbf{U}\mathbf{1})$ (where
$\mbox{diag}(\mathbf{a})$ is the diagonal matrix having as main diagonal
vector $\mathbf{a}$ and $\mathbf{1}$ is a vector of ones). The objective
function with the chosen
weights $w_{ij}$ results in a heavy penalty if the neighbouring points
$\mathbf{x}_i$ and $\mathbf{x}_j$ are mapped far apart. Therefore,
its minimization ensures that if $\mathbf{x}_i$ and $\mathbf{x}_j$ are
near,
then the projected features $\mathbf{y}_i = \mathbf{W}^T \mathbf{x}_i$
and $\mathbf{y}_j=\mathbf{W}^T\mathbf{x}_i$ are near, as well.  To the best of our knowledge no probabilistic models exist for LPPs. In
the following (Sec. \ref{sec:ML}, \ref{sec:aunifiedEMforCA}), we show how a probabilistic version of LPPs arises by
choosing an appropriate prior over
the latent space $\mathbf{y}_i$.

\subsection{Slow Feature Analysis}

Now let us consider the case that the columns of $\mathbf{x}_i$ are
samples of a time series of length $T$.
The aim of Slow Feature Analysis (SFA) is, given $T$ sequential observation
vectors $\mathbf{X} = [\mathbf{x}_1 \ldots \mathbf{x}_T]$, to find an
output signal representation $\mathbf{Y} = [\mathbf{y}_1 \ldots
\mathbf{y}_T]$ for which the features change slowest over time
\cite{wiskott2002slow}, \cite{klampfl2009replacing}. By assuming again
a linear mapping $\mathbf{Y} = \mathbf{W}^T \mathbf{X}$ for the output
representation, SFA minimizes the {\it slowness} for these values, defined as
the variance of
the first derivative of $\mathbf{Y}$. Formally, $\mathbf{W}$ of SFA
is computed as
\begin{equation}\label{E:SFA}
\mathbf{W}_o = \arg\min_{\mathbf{W}} \mbox{tr}\left[ \mathbf{W}^T
\mathbf{\dot{X}} \mathbf{\dot{X}} \mathbf{W} \right], \,\,
\mbox{s.t.}\,\, \mathbf{W}^T\mathbf{S}\mathbf{W} = \mathbf{I},
\end{equation}
where $\mathbf{\dot{X}}$ is the first derivative matrix (usually computed as the
first order difference i.e., $\mathbf{\dot{x}}_j = \mathbf{x}_j -
\mathbf{x}_{j-1}$).  An ML solution of SFA was recently proposed in
\cite{SFA}, by incorporating a Gaussian linear
dynamic system prior over the latent space $\mathbf{Y}$. The proposed
generative model is
\begin{equation}
\begin{array}{rl}
P(\mathbf{x}_t|\mathbf{W},\mathbf{y}_t,\sigma_x)&=\mathcal{N}(\mathbf{W}^{-1}\mathbf{y}_t,\sigma_x^2\mathbf{I})\\
P(\mathbf{y}_t|\mathbf{y}_{t-1},\lambda_{1:N},\sigma_{1:N})&=
\prod_{n=1}^N P(y_{n,t}|y_{n,t-1},\lambda_n,\sigma_n^2)
\end{array}
\end{equation}
with $P(y_{n,t}|y_{n,t-1},\lambda_n,\sigma_n^2)=\mathcal{N}(\lambda_n y_{n,t-1},\sigma_n^2)$ and $P(y_{n,1}|\sigma_{n,1}^2)=\mathcal{N}\left(0,\sigma_{n,1}^2\right)$.
As we will show, SFA is indeed a special case of our general model.

\looseness-1Summarizing, in the following sections we formulate a unified,
probabilistic framework for component analysis which: (1) incorporates PCA as a special case, (2) produces a Probabilistic LDA which (i) has an ML solution for the loading matrix $\mathbf{W}$ with similar direction to the deterministic LDA (Eq. \ref{eq:ldaopt}) and (ii) does not make assumptions regarding the number of samples per class (as in \cite{IoffeECCV2006}), (3) provides the first, to the  best of our knowledge, probabilistic model that explains LPP, (4) naturally incorporates SFA as a special case, (5) provides variance estimates not only for observations but also per latent dimension (differentiating our approach from existing probabilistic CA (e.g., PPCA, PLDA), and (6) provides a straightforward framework for producing novel component analysis techniques.

\section{A Unified ML Framework for Component Analysis}
\label{sec:ML}

In this section, we will present the proposed Maximum Likelihood (ML) framework for probabilistic component analysis and show how PCA, LDA, LPP and SFA can be generated within this framework, also proving equivalence with known deterministic models.  Firstly, to ease computation, we assume the generative model for the $i$-th observation, $\mbf{x}_i$, is defined as
\begin{equation}
\mathbf{x}_i = \mathbf{W}^{-1}\mathbf{y}_i + \mbox{\boldmath$\epsilon$}_{i},\,\, \mbox{\boldmath$\epsilon$}_{i} \sim N(\mathbf{0},\sigma_x^2\mathbf{I}).\label{eq:ML_generative_model}
\end{equation}
In order to fully define the likelihood we need to define a prior distribution on the latent variables $\mbf{y}$.  We will prove that by choosing one of the priors defined below and subsequently taking the ML solution wrt. parameters, we end up generating the aforementioned family of probabilistic component models. The priors, parametrised by $\beta=\{\sigma_{1:N},\lambda_{1:N}\}$, are as follows (see also Fig. \ref{fig:motivating}).

$\bullet$ An MRF with full connectivity - each latent node $\mbf{y}_i$ is connected to all other latent nodes $\mbf{y}_j, j \neq i$.
        \begin{equation}
        \begin{array}{rl}
         P(\mathbf{Y}|\beta) &=\frac{1}{Z} \exp\left\{ -\frac{1}{2}\sum_{n=1}^N \sum_{i=1}^T \frac{1}{T-1}\sum_{j=1,j\neq i}^T \frac{1}{\sigma_n^2}(y_{n,i} - \lambda_n y_{n,j})^2 \right \}\\
&\approx\frac{1}{Z} \exp\left\{ -\frac{1}{2}\sum_{n=1}^N \sum_{i=1}^T \frac{1}{T}\sum_{j=1}^T \frac{1}{\sigma_n^2}(y_{n,i} - \lambda_n y_{n,j})^2 \right \}\\         
                       &= \frac{1}{Z} \exp\left\{ -\frac{1}{2} \left(\mbox{tr}\left[   \mathbf{\Lambda}^{(1)}\mathbf{Y}\mathbf{Y}^T\right ] +  \mbox{tr}\left[   \mathbf{\Lambda}^{(2)}\mathbf{Y}\mathbf{M}\mathbf{Y}^T\right ]  \right) \right\},
         \end{array}\label{eq:prior:PCA}
         \end{equation}
         where $\mathbf{M} \triangleq -\frac{1}{T} \mathbf{1}\mathbf{1}^T$, $\mathbf{\Lambda}^{(1)}\triangleq \left[\delta_{mn}\frac{\lambda_n^2 +1}{\sigma_n^2}\right],\mathbf{\Lambda}^{(2)}\triangleq \left[\delta_{mn}\frac{\lambda_n}{\sigma_n^2}\right]$.

$\bullet$ A product of two MRFs. In the first, each latent node $\mbf{y}_i$ is connected only to other latent nodes in the same class ($\mbf{y}_j, j \in \mathcal{\tilde{C}}_i$).  In the second, each latent node ($\mbf{y}_i$) is connected to all other latent nodes ($\mbf{y}_j, j\neq i$).
     \begin{equation}
     \begin{array}{rl}
    P(\mathbf{Y}|\beta)=&\frac{1}{Z} \exp \left\{-\frac{1}{2}\sum_{n=1}^N \sum_{i=1}^T \frac{1}{|{\tilde{\mathcal{C}_i}}|} \sum_{j \in \tilde{\mathcal{C}}_i} \frac{\lambda_n}{\sigma_n^2} (y_{n,i} - y_{n,j})^2  \right \}\\
     &\exp \Big\{-\frac{1}{2}\sum_{n=1}^N \sum_{i=1}^T \frac{1}{T-1} \sum_{j=1}^T 
     \frac{(1-\lambda_n)^2}{\sigma_n^2}(y_{n,i} - y_{n,j})^2 \Big \}\\
      =& \frac{1}{Z} \exp \left\{ -\frac{1}{2}\left ( \mbox{tr}\left[   \mathbf{\Lambda}^{(1)}\mathbf{Y}\mathbf{M}_{c}\mathbf{Y}^T \right ] + \mbox{tr}\left[ \mathbf{\Lambda}^{(2)}\mathbf{Y}\mathbf{M}_t\mathbf{Y}^T \right ]    \right ) \right\},
     \end{array}\label{eq:prior:LDA}
     \end{equation}
      where $\mathbf{M}_c \triangleq \mathbf{I} - \mbox{diag}[\mathbf{C}_1,\ldots,\mathbf{C}_C]$, $\mathbf{C}_c \triangleq \frac{1}{|\mathcal{{C}}_c|} \mathbf{1}_c\mathbf{1}^T_c$,
      $\mathbf{M}_t \triangleq \mathbf{I} + \mathbf{M}$, $\mathbf{\Lambda}^{(1)} \triangleq \left[\delta_{mn}(\frac{\lambda_n}{\sigma_n^2})\right]$, $\mathbf{\Lambda}^{(2)} \triangleq \left[\delta_{mn}\frac{(1-\lambda_n)^2}{\sigma_n^2}\right]$, while $\tilde{\mathcal{C}}_i=\{j : \exists\;\mathcal{C}_l\;\textrm{s.t.}\;\{\mbf{x}_j,\mbf{x}_i\} \in \mathcal{C}_l, i\neq j\}$. 

$\bullet$ A product of two MRFs. In the first, each latent node $\mbf{y}_i$ is connected to all other latent nodes that belong in $\mbf{y}_i$'s neighbourhood (symmetrically defined as $\mathcal{N}_{i}^s=\mathcal{N}_{j}^s=\{i \in \mathcal{N}_j\cup j \in \mathcal{N}_i\}$).  
In the second, we only have individual potentials per node.
\begin{equation}
     \begin{array}{rl}
     P(\mathbf{Y}|\beta)  =& \frac{1}{Z} \exp \Big(-\frac{1}{2}\sum_{n=1}^N \sum_{i=1}^T \frac{1}{|{\mathcal{N}_i^s}|}\sum_{j \in \mathcal{N}_i^s} \frac{\lambda_n}{\sigma_n^2}(y_{n,i} - y_{n,j})^2  \Big)\\&
     \exp \Big(-\frac{1}{2}\sum_{n=1}^N\sum_{i=1}^T \frac{(1-\lambda_n)^2}{\sigma_n^2}y_{n,i}^2 \Big ) \\
      =&  \frac{1}{Z} \exp \left\{ -\frac{1}{2}\left ( \mbox{tr}\left[   \mathbf{\Lambda}^{(1)}\mathbf{Y}\mathbf{\tilde{L}}\mathbf{Y}^T \right ] + \mbox{tr}\left[ \mathbf{\Lambda}^{(2)}\mathbf{Y}\mathbf{\tilde{D}}\mathbf{Y}^T \right ]    \right ) \right\}
     \end{array}\label{eq:prior:LE}
     \end{equation}
where  $\mathbf{\tilde{L}}=\mathbf{D}^{-1}\mathbf{L}$ and $\mathbf{\tilde{D}}=\mbf{I}$ ($\mathbf{{L}}$ and $\mathbf{{D}}$ are defined in Sec. \ref{S:LPP} referring to LPPs).  $\mathbf{\Lambda}^{(1)}$ and $\mathbf{\Lambda}^{(2)}$ are defined as above.

$\bullet$ A linear dynamical system prior over the latent space.
\begin{equation}
     \begin{array}{rl}
      P(\mathbf{Y}|\beta) &= \frac{1}{Z}\exp\Big\{-\sum_{n=1}^N\Big(\frac{1}{2\sigma_{n,1}^2}y_{n,1}^2+\frac{1}{2\sigma_{n}^2}\sum_{t=2}^T[y_{n,t}-\lambda_ny_{n,t-1}]^2\Big)\Big\} \\
                    &\approx \frac{1}{Z} \exp \left\{ -\frac{1}{2}\left ( \mbox{tr}\left[   \mathbf{\Lambda}^{(1)}\mathbf{Y}\mathbf{K_1}\mathbf{Y}^T \right ] + \mbox{tr}\left[ \mathbf{\Lambda}^{(2)}\mathbf{Y}\mathbf{Y}^T \right ]    \right ) \right\}
          \end{array}\label{eq:prior:SFA}
     \end{equation}
     where $\mathbf{K}_1 = \mathbf{P}_1\mathbf{P}_1^T $ and $\mathbf{P}_1$ is a $T\times(T-1)$ matrix with elements $p_{ii} = 1 $ and $p_{(i+1)i}=-1$ (the rest are zero).  The approximation holds when $T \rightarrow \infty$.  Again, $\mathbf{\Lambda}^{(1)}$ and $\mathbf{\Lambda}^{(2)}$ are defined as above.

\looseness-1In all cases the partition function $Z$ is defined as $Z =\int P(\mathbf{Y}) d \mathbf{Y}$.  The motivation behind choosing the above latent priors was given by the influential analysis made in \cite{he2005face} where the connection between (deterministic) LPP, PCA and LDA was explored.  A further piece of the puzzle was added by the recent work \cite{SFA} where the linear dynamical system prior (Eq. \ref{eq:prior:SFA}) was used in order to provide a derivation of SFA in a ML framework.
By formulating the appropriate priors for these models we unify these subspace methods in a single probabilistic framework of a linear generative model along with a prior of the form 
\hspace*{-0.2pt}
\begin{equation}
\begin{array}{l}
P(\mathbf{Y})\propto\exp\left\{ -\frac{1}{2} \left(\mbox{tr}\left[   \mathbf{\Lambda}^{(1)}\mathbf{{Y}}\mathbf{B}^{(1)}\mathbf{{Y}}^T \right ] + \mbox{tr}\left[ \mathbf{\Lambda}^{(1)}\mbf{{Y}}\mathbf{B}^{(2)}\mathbf{{Y}}^T \right ] \right )\right \}.
\end{array}
\label{eq:general_prior}
\end{equation}
\looseness-1The differentiation amongst these models lies in the neighbourhood over which the
potentials are defined.  In fact, the varying neighbouring system is translated into the matrices $\mathbf{B}^{(1)}$ and $\mathbf{B}^{(2)}$ in the functional form of the potentials, essentially encapsulating the latent covariance connectivity.  E.g., for Eq. \ref{eq:prior:PCA}, $\mathbf{B}^{(1)}=\mbf{I}$ and $\mathbf{B}^{(2)}=\mbf{M}$, for Eq. \ref{eq:prior:LDA}, $\mathbf{B}^{(1)}=\mbf{M}_c$ and $\mathbf{B}^{(2)}=\mbf{M}_t$, for Eq. \ref{eq:prior:LE}, $\mathbf{B}^{(1)}=\mbf{\tilde{L}}$ and $\mathbf{B}^{(2)}=\mbf{\tilde{D}}$ and finally for Eq. \ref{eq:prior:SFA}, $\mathbf{B}^{(1)}=\mbf{K}$ and $\mathbf{B}^{(2)}=\mbf{I}$. 
\looseness-1In the following we will show that ML estimation using these potentials is equivalent to the deterministic formulations of PCA, LDA and LPP.  SFA is a special case for which it was already shown in \cite{SFA} that a potential of the form of Eq. \ref{eq:prior:SFA} with an ML framework produces a projection with the same direction as Eq. \ref{E:SFA}.

Adopting the linear generative model in Eq. \ref{eq:ML_generative_model}, the corresponding conditional data (observation) probability is a Gaussian,
\begin{equation}\label{E:GEN}
P(\mathbf{x}_t|\mathbf{y}_t,\mathbf{W},\sigma_x^2) = \mathcal{N}(\mathbf{W}^{-1}\mathbf{y}_t,\sigma_x^2).
\end{equation}
Having chosen a prior of the form described in Eq. \ref{eq:general_prior} (e.g., as defined in  Eq. \ref{eq:prior:PCA},\ref{eq:prior:LDA},\ref{eq:prior:LE},\ref{eq:prior:SFA}) we can now derive the likelihood of our model as follows:
\begin{equation}
P(\mathbf{X}|\Psi ) = \int {\prod_{t=1}^T {P(\mathbf{x}_t|\mathbf{y}_t,\mathbf{W},\sigma^2})P(\mathbf{Y}|\sigma^2_{1:N},\lambda_{1:N})d \mathbf{Y},
\label{eq:model}} \end{equation}
where the model parameters are defined as $\Psi=\{\sigma_x^2, \mathbf{W},\sigma^2_{1:N},\lambda_{1:N}\}$.  In the following we will show that by substituting the above priors in Eq. \ref{eq:model} and maximising the likelihood we obtain loadings $\mathbf{W}$ which are co-directional (up to a scale ambiguity) to deterministic PCA, LDA and LPPs and SFA. 
Firstly, by substituting the general prior (Eq. \ref{eq:general_prior}) in the likelihood (Eq. \ref{eq:model}), we obtain
\begin{equation}
\begin{array}{rl}
P(\mathbf{X}|\Psi )&=\int {\prod_{t=1}^T P(\mathbf{x}_t|\mathbf{y}_t,\mathbf{W},\sigma^2})\frac{1}{Z} \exp\\
&\left\{ -\frac{1}{2} \left(\mbox{tr}\left[   \mathbf{\Lambda}^{(1)}\mathbf{Y}\mathbf{B}^{(1)}\mathbf{Y}^T\right ] + \mbox{tr}\left[   \mathbf{\Lambda}^{(2)}\mathbf{Y}\mathbf{B}^{(2)}\mathbf{Y}^T\right ]  \right) \right\} d\mathbf{Y}.
\end{array}\label{eq:model2}\end{equation}
In order to obtain a zero-variance limit ML solution, we map $\sigma_x\rightarrow 0$
\begin{equation}
\begin{array}{rl}
P(\mathbf{X}|\Psi ) &= \int \prod_{t=1}^T{\delta(\mathbf{x}_t-\mathbf{W}^{-1}\mathbf{y}_t)}\frac{1}{Z} \exp \\
&\left\{ -\frac{1}{2} \left(\mbox{tr}\left[   \mathbf{\Lambda}^{(1)}\mathbf{{Y}}\mathbf{B}^{(1)}\mathbf{{Y}}^T\right ] + \mbox{tr}\left[   \mathbf{\Lambda}^{(2)}\mathbf{{Y}}\mathbf{B}^{(2)}{Y}^T\right ]  \right) \right\} d\mathbf{Y}
\label{eq:model2}\end{array} \end{equation}
By completing the integrals and taking the $\log$, we obtain the conditional log-likelihood:
\begin{equation}\label{E:LOGLIKE}
\begin{array}{rl}
L(\Psi)=\log P(\mathbf{X}|\theta )=&-\log Z + T\log|\mathbf{W}|-\frac{1}{2}\\
&\mbox{tr} \left [\mathbf{\Lambda}^{(1)}\mathbf{W}\mathbf{X}\mathbf{B}^{(1)}\mathbf{X}^T\mathbf{W}^T  +  \mathbf{\Lambda}^{(2)}\mathbf{W}\mathbf{X}\mathbf{B}^{(2)}\mathbf{X}^T\mathbf{W}^T   \right ]
\end{array}
\end{equation}
where $Z $ is a constant term independent of $\mathbf{W}$. By maximising for $\mathbf{W}$ we obtain
\begin{equation}
\begin{array}{r}
T  \mathbf{W}^{-T} - \left(\mathbf{\Lambda}^{(1)}\mathbf{W}\mathbf{X}\mathbf{B}^{(1)}\mathbf{X}^T +  \mathbf{\Lambda}^{(2)}\mathbf{W}\mathbf{X}\mathbf{B}^{(2)}\mathbf{X}^T\right) = \mathbf{0}, \\
 \mathbf{I} = \mathbf{\Lambda}^{(1)}\mathbf{W}\mathbf{X}\mathbf{B}^{(1)}\mathbf{X}^T\mathbf{W}^T +  \mathbf{\Lambda}^{(2)}\mathbf{W}\mathbf{X}\mathbf{B}^{(2)}\mathbf{X}^T\mathbf{W}^T.
\end{array}\label{eq:finalll}
\end{equation}
\looseness-1It is easy to prove that since $\mathbf{\Lambda}^{(1)},\mathbf{\Lambda}^{(2)}$ are diagonal matrices, the $\mathbf{W}$ which satisfies Eq. \ref{eq:finalll} simultaneously diagonalises (up to a
scale ambiguity) $\mathbf{X}\mathbf{B}^{(1)}\mathbf{X}^T$  and $\mathbf{X}\mathbf{B}^{(2)}\mathbf{X}^T$.  By substituting the $\mathbf{B}$ matrices as defined above in Eq. \ref{eq:finalll}, we now consider all cases separately.  For PCA, by utilising Eq. \ref{eq:prior:PCA}, Eq. \ref{eq:finalll} is reformulated as $\mathbf{W}\mathbf{X}\mathbf{X}^T \mathbf{W}^T = \left[\mathbf{\Lambda}^{(1)}\right]^{-1}$ hence
 $\mathbf{W}$ is given by the eigenvectors of the total scatter matrix $\mathbf{S}$.  For LDA (Eq. \ref{eq:prior:LDA}), Eq. \ref{eq:finalll} is reformulated as $\mathbf{\Lambda}^{(1)}\mathbf{W}\mathbf{X}\mathbf{M}\mathbf{X}^T\mathbf{W}^T +  \mathbf{\Lambda}^{(2)}\mathbf{W}\mathbf{X}\mathbf{X}^T\mathbf{W}^T = \mathbf{I}$.  Thus, $\mathbf{W}$  is given by the directions that simultaneously diagonalise $\mathbf{S}$ and $\mathbf{S}_w$.  For LPP (Eq. \ref{eq:prior:LE} ), Eq. \ref{eq:finalll} yields $\mathbf{\Lambda}^{(1)}\mathbf{W}\mathbf{X}\mathbf{L}\mathbf{X}^T \mathbf{W}^T+  \mathbf{\Lambda}^{(2)}\mathbf{W}\mathbf{X}\mathbf{D}^T\mathbf{X}^T \mathbf{W}^T= \mathbf{I} $, therefore $\mathbf{W}$  is given by the directions that simultaneously diagonalise  $\mathbf{X}\mathbf{\tilde{L}}\mathbf{X}^T$ and $\mathbf{X}\mathbf{\tilde{D}}\mathbf{X}^T$.  Finally, for SFA, by utilising Eq. \ref{eq:prior:SFA}, Eq. \ref{eq:finalll} becomes $\mathbf{\Lambda}^{(1)}\mathbf{W}\mathbf{X}\mathbf{K}\mathbf{X}^T\mathbf{W}^T +  \mathbf{\Lambda}^{(2)}\mathbf{W}\mathbf{X}\mathbf{X}^T\mathbf{W}^T = \mathbf{I}$, and  $\mathbf{W}$   is given by the directions that simultaneously diagonalise $\mathbf{X}\mathbf{K}\mathbf{X}^T$ and $\mathbf{X}\mathbf{X}^T$.

The above shows that the ML solution following our framework is equivalent to the deterministic models of PCA, LDA, LPP and SFA.  The direction of $\mathbf{W}$ does not depend of $\sigma_n^2$ and $\lambda_n$, which can be estimated by optimizing Eq. \ref{E:LOGLIKE} with regards to these parameters. In this work we will provide update rules for $\sigma_n$ and $\lambda_n$ using an EM framework. As we observe, the ML loading $\mathbf{W}$ does not depend on the exact setting of
$\lambda_n$, so long as they are all different. If $0 < \lambda_n < 1,\, \forall\,\, n$, then larger values of $\lambda_n$
correspond to more expressive (in case of PCA), more discriminant (in LDA), more local (in LPP) and slower latents (in case of SFA). 
This corresponds directly to the ordering of the solutions from PCA, LDA, LPP and SFA. To recover exact equivalence to LDA, LPP, SFA another limit is required that corrects the scales.
There are several choices, but a natural one is to let $\sigma_n^2 = 1-\lambda_n^2$. This choice in case of LDA and SFA fixes the prior
covariance of the latent variables to be one ($\mathbf{W}^T\mathbf{X}\mathbf{X}\mathbf{W} = \mathbf{I}$) and it forces $\mathbf{W}^T\mathbf{X}\mathbf{D}\mathbf{X}\mathbf{W} = \mathbf{I}$
in case of LPP.  This choice of $\sigma_n$ has been also discussed in \cite{SFA} for SFA.  We note that in case of PCA, we should set $\sigma_n$ to be analogous to the corresponding eigenvalue of the covariance matrix, since otherwise the method will result to a {\it minor} component analysis.

\section{A Unified EM Framework for Component Analysis }
\label{sec:aunifiedEMforCA}
\def\uniFirst{1} 

In the following we propose a unified EM framework for component analysis. This framework can treat all priors with undirected links (such as Eq. \ref{eq:prior:PCA}, Eq. \ref{eq:prior:LDA} and Eq. \ref{eq:prior:LE}).  The EM of the prior in Eq. \ref{eq:prior:SFA}  contains only  directed links with no loops, and thus can be solved (without any approximations) similarly to the EM of a linear dynamical system \cite{BishopBook}.  If we treat the SFA links as undirected, we end up with an autoregressive component analysis (see Section \ref{sec:EM_discussion}). 

In order to perform EM with an MRF prior we adopt the simple and elegant mean field approximation theory \cite{qian1991estimation,celeux2003procedures,zhang1992mean}, which essentially allows computationally favourable factorizations within an EM framework.
\if\uniFirst1  
Let us consider a generalisation of the priors we defined in Sec. \ref{sec:ML} to $\mathcal{M}$ MRFs:
\begin{gather}
P(\mbf{Y}|\beta)=\prod_{\mu \in \mathcal{M}}\frac{1}{Z^\mu}\exp\left\{Q^\mu\right\}\label{eq:genMultipleMRFprior}\\
Q^\mu=-\sum_{n=1}^N\frac{f_\mu(\lambda_n)}{2\sigma_n^2}\frac{1}{c}\sum_{i \in \omega_i}\frac{1}{c_j^\mu}\sum_{j \in \omega_j^\mu}(y_{n,i}-\phi_\mu(\lambda_n)y_{n,j})^2\nonumber
\end{gather}
\looseness-1where $c$ and $c_j$ are normalisation constants, while $f_\mu$ and $\phi_\mu$ are  functions of $\lambda_n$.  Without loss of generality and for clarity of notation, we assume that $c=1$, $c_j^\mu=|\omega_j^\mu|$ and $\omega_i^\mu=[1,\dots,T]$.  Furthermore, we now assume the linear model 
\else
Let us consider the priors we defined in Sec. \ref{sec:ML}.  Without loss of generality, we now assume the linear model 
\fi
\begin{equation}
\mbf{x}_i=\mbf{W}\mbf{y}_i+\epsilon_i, \epsilon_i \sim \mathcal{N}(0,\sigma_x^2).
\label{eq:EM:em_model}
\end{equation}
\looseness-1For clarity, the set of parameters associated with the prior (i.e. energy function) are denoted as $\beta=\{\sigma_{1:N}, \lambda_{1:N}\}$,  the parameters related to the observation model $\theta=\{\mbf{W},\sigma_x\}$, while the total parameter set is denoted as $\Psi=\{\theta, \beta\}$. 
In agreement with \cite{celeux2003procedures}, we replace the marginal distribution $P(\mbf{Y}|\beta)$ by the mean-field
\if\uniFirst1  
\begin{equation}
P(\mbf{Y}|\beta)\approx \prod_{i=1}^TP(\mbf{y}_i|\mbf{m}_i^\mathcal{M},\beta^\mathcal{M}).
\end{equation}
\else
\begin{equation}
P(\mbf{Y}|\beta)\approx \prod_{i=1}^TP(\mbf{y}_i|\mbf{m}_i^{(R)},\beta^\mathcal{M}).
\end{equation}
\fi
Since different CA  models have different latent connectivities (and thus different MRF configurations), the mean-field influence on each latent point $\mbf{y}_i$ now depends on the model-specific connectivity via $\mbf{m}_i^{\mathcal{M}}$, a function of $\mathbb{E}[\mbf{y}_j]$.
After calculating the normalising integral for the priors Eq. \ref{eq:prior:PCA}-\ref{eq:prior:LE}  and given the mean-field, 
\if\uniFirst1
it can be easily shown that Eq. \ref{eq:genMultipleMRFprior} follows a Gaussian distribution,
\begin{equation}
P(\mathbf{y}_i|\mathbf{m}_{i}^{\mathcal{M}},\beta)=\mathcal{N}(\mbf{m}_i^\mathcal{M},\mbf{\Sigma}^\mathcal{M}),
\end{equation}
\begin{gather}
\mbf{m}_i^\mathcal{M}=\sum_{\mu \in \mathcal{M}}\left(\frac{f_{\mu}(\lambda_n)\phi_\mu(\lambda_n)}{F^M(\lambda_n)}\boldsymbol\mu_{\omega_j^\mu}\right)=\sum_{\mu \in \mathcal{M}}\mathbf{\Lambda}^{{\mu}}\boldsymbol\mu_{\omega_j^\mu}\label{eq:genMultipleMRFpriorMean},\\
\mbf{\Sigma}^\mathcal{M}=\left[\delta_{mn}\frac{\sigma_n^2}{F^M(\lambda_n)}\right]\label{eq:genMultipleMRFpriorVar}
\end{gather}
with $\boldsymbol\mu_{\omega_j^\mu}=\frac{1}{|\omega_j^\mu|}\sum_{j\in \omega_j^\mu}\mathbb{E}[\mbf{y}_{n,j}]$ and $F^M(\lambda_n)=\sum_{\mu \in \mathcal{M}}f_\mu(\lambda_n)$. 
\else
we obtain
\fi

\begin{table}[htbp]
  \centering
\def\arraystretch{0.5}
\setlength{\tabcolsep}{3pt}  
  \caption{MRF configuration for PCA, LDA and LPP.}
    \begin{tabular}{c|c|ccc|ccc}
    \toprule
    {$\mathcal{M}=\{\alpha,\beta\}$}     & $F^\mathcal{M}=\sum_\mu f_\mu$ &$f_a$ & $\phi_\alpha$ & $\omega_j^\alpha$ & $f_\beta$ & $\phi_\beta$ & $\omega_j^\beta$ \\
    \midrule
       PCA (\ref{eq:prior:PCA})  & 1& 1     & $\lambda_n$ &  {$\{1\dots T\}\setminus \{i\}$} &       &       &  \\
    LDA (\ref{eq:prior:LDA})  &{$\lambda_n+(1-\lambda_n)^2$}& $\lambda_n$ & 1     & $\tilde{\mathcal{C}}_i$     & {$(1-\lambda_n)^2$} & 1     & $\{1\dots T\}\setminus \{i\}$ \\
LPP  (\ref{eq:prior:LE}) &{$\lambda_n+(1-\lambda_n)^2$}& $\lambda_n$ & 1     & $\mathcal{N}_i^s$     & {$(1-\lambda_n)^2$} & 0     & $\{1\}$\\
    \bottomrule
    \end{tabular}
  \label{tab:generalMRFconfiguration}
\end{table}
\if\uniFirst1
Therefore, by simply replacing the parametrisation of the priors we defined in Eq. \ref{eq:prior:PCA} (PCA), \ref{eq:prior:LDA} (LDA) and \ref{eq:prior:LE} (LPP) (see also Tab. \ref{tab:generalMRFconfiguration}) for the mean and variance (Eq. \ref{eq:genMultipleMRFpriorMean} and Eq. \ref{eq:genMultipleMRFpriorVar}), we obtain the distribution for each CA method we propose.
\else
\begin{equation}
\begin{array}{rl}
P(\mathbf{y}_i|\mathbf{m}_{i}^{\mathcal{M}},\beta)& = 
 \left\{
\begin{array}{l}
\mathcal{N}(\mathbf{y}_i| \mathbf{m}_i^{\textrm{(PCA)}},\mathbf{\Sigma}^{\textrm{(PCA)}})\,\mbox{for}\,(\ref{eq:prior:PCA})  \\
\mathcal{N}(\mathbf{y}_i| \mathbf{m}_i^{\textrm{(LDA)}}, \mathbf{\Sigma}^{\textrm{(LDA)}})\, \mbox{for}\,(\ref{eq:prior:LDA}) \\
\mathcal{N}(\mathbf{y}_i| \mathbf{m}_i^{\textrm{(LPP)}}, \mathbf{\Sigma}^{\textrm{(LPP)}})\, \mbox{for}\,(\ref{eq:prior:LE})
\end{array}
\right.
\\
\end{array}\label{eq:meanfield}
\end{equation}
where  $R=\{\mbox{PCA},\mbox{LDA},\mbox{LPP}\}$.  
\fi
The means $\mbf{m}_i^{\mathcal{M}}$ for PCA, LDA and LPP are obtained as
\begin{equation}
\begin{array}{l}
\mbf{m}_{i}^{\textrm{(PCA)}} = \mathbf{\Lambda}\boldsymbol\mu_{-i},
\mbf{m}_{i}^{\textrm{(LDA)}} = \mathbf{\Lambda}^{(\alpha)}\boldsymbol\mu_{-i} + \mathbf{\Lambda}^{(\beta)}\boldsymbol\mu_{\tilde{\mathcal{C}}_i},
\mbf{m}_{i}^{\textrm{(LPP)}} = \mathbf{\Lambda}^{(\alpha)}\boldsymbol\mu_{\mathcal{N}_i^s}
\end{array}
\label{eq:EM:means_mi}
\end{equation}
and the variances $\bsym\Sigma^{\mathcal{M}}$  as 
\begin{equation}
\begin{array}{l}
\boldsymbol\Sigma^{\textrm{(PCA)}} = \left[\delta_{mn}\sigma_n^2\right],
\boldsymbol\Sigma^{\textrm{(LDA)}}=\boldsymbol\Sigma^{\textrm{(LPP)}}=\left[\delta_{mn}\left(\frac{\sigma_n^2}{\lambda_n + (1 - \lambda_n)^2}\right)\right]
\label{eq:EM:variances_S}
\end{array}
\end{equation}
where $\boldsymbol\mu_{-i} = \frac{1}{T-1}\sum_{j \neq i}^T \mathbb{E}^{\mathcal{M}}[\mbf{y}_j]$ is the mean, $\boldsymbol\mu_{\tilde{\mathcal{C}}_i}=\frac{1}{|{\tilde{\mathcal{C}}_i}|}\sum_{j \in \tilde{\mathcal{C}}_i} \mathbb{E}^{\mathcal{M}}[\mbf{y}_j]$ the class mean, and $\boldsymbol\mu_{\mathcal{N}_i^s}=\frac{1}{|{\mathcal{N}_i^s}|}\sum_{j \in \mathcal{N}_i^s}^T \mathbb{E}^{\mathcal{M}}[\mbf{y}_j]$ the neighbourhood mean. Furthermore, $\mathbf{\Lambda} = \left[\delta_{mn}\lambda_n\right]$, $\mathbf{\Lambda}^{(\alpha)}=\left[\delta_{mn}\left(\frac{\lambda_n}{\lambda_n + (1 - \lambda_n)^2}\right)\right]$ and $\mathbf{\Lambda}^{(\beta)}=\left[\delta_{mn}\left(\frac{(1-\lambda_n)^2}{\lambda_n + (1 - \lambda_n)^2}\right)\right]$.

In order to complete the expectation step, we infer the first order moments of the latent posterior, defined as
\begin{equation}
P(\mbf{y}_i|\mbf{x}_i,\mbf{m}_i^{\mathcal{M}},\Psi^{\mathcal{M}})=\frac{P(\mathbf{x}_i|\mathbf{y}_i,\theta^{\mathcal{M}})P(\mathbf{y}_i|\mathbf{m}_i^{\mathcal{M}},\beta^{\mathcal{M}})}{\int_{\mathbf{y}_i}P(\mathbf{x}_i|\mathbf{y}_i,\theta^{\mathcal{M}})P(\mathbf{y}_i|\mathbf{m}_i^{\mathcal{M}},\beta^{\mathcal{M}})d\mathbf{y}_i}.\label{eq:latentposterior}
\end{equation}
Since the posterior is a product of Gaussians\footnote{The result can be easily obtained by completing the square for $\mbf{y}_i$.}
, we have
\begin{equation}
P(\mbf{y}_i|\mbf{x}_i,\mbf{m}_i^{\mathcal{M}},\Psi^{\mathcal{M}})=\mathcal{N}(\mbf{y}_i|(\mbf{W}^T\mbf{x}_i+{\boldsymbol\Sigma}^{\mathcal{M}^{-1}}\mbf{m}_i^{\mathcal{M}})\mbf{A},\sigma_x^{\mathcal{M}^2}\mbf{A})\label{eq:the_latent_posterior}
\end{equation}
with $\mbf{A}=(\mathbf{W}^T\mathbf{W}+(\hat{\mathbf{{\Sigma}}}^{\mathcal{M}})^{-1})^{-1}$ and $\hat{\boldsymbol\Sigma}^{\mathcal{M}}=\left[\delta_{mn}({\Sigma_{mn}^{\mathcal{M}}} / {\sigma_x^{\mathcal{M}^2}})\right]$.  Therefore $\mathbb{E}^{\mathcal{M}}[\mbf{y}_i]$ is equal to the mean, and $\mathbb{E}^{\mathcal{M}}[\mbf{y}_i\mbf{y}_i^T]=\sigma_x^{\mathcal{M}^2}\mbf{A} + \mathbb{E}[\mbf{y}_i]\mathbb{E}[\mbf{y}_i]^T$.

Having recovered the first order moments, we move on to the maximisation step.  In order to maximize the marginal log-likelihood, $\log{P(\mbf{X}|\Psi^\mathcal{M})}$, we adopt the usual EM bound \cite{RoweisZoubin}, $\int_{\mbf{Y}}P(\mbf{Y}|\mbf{X},\Psi^\mathcal{M})\log{P(\mbf{X},\mbf{Y})}d\mbf{Y}$.  By adopting the approximation proposed in \cite{celeux2003procedures}, the complete-data likelihood is factorised as
\begin{equation}
P(\mbf{Y},\mbf{X}|\Psi^{\mathcal{M}})\approx\prod_{i=1}^T P(\mbf{x}_i|\mbf{y}_i,\theta^{\mathcal{M}})P(\mbf{y}_i|\mbf{m}_i^{\mathcal{M}},\beta^{\mathcal{M}}).
\end{equation}
and therefore, the maximisation term (EM bound) becomes
\begin{equation}\begin{array}{r}
\sum_{i=1}^T\int_{\mathbf{y}_i} P(\mathbf{y}_i|\mbf{x}_i,\mathbf{m}_i^{\mathcal{M}},\Psi^{\mathcal{M}})\log P(\mathbf{x}_i,\mathbf{y}_i|\Psi^{\mathcal{M}})d\mathbf{y}_i.\label{eq:maxall}
\end{array}\end{equation}
As can be seen the likelihood can be separated due to the logarithm for estimating $\theta^{\mathcal{M}}=\{\mathbf{W}^{\mathcal{M}}, \sigma_x^{\mathcal{M}}\}$ and $\beta=\{\sigma_{1:N}^{\mathcal{M}},\lambda_{1:N}^{\mathcal{M}}\}$ as follows:
\begin{equation}\begin{array}{l}
\theta^{\mathcal{M}}= \arg\max\Big\{\sum_{i=1}^T\int_{\mathbf{y}_i} P(\mathbf{y}_i|\mbf{x}_i,\mathbf{m}_i^{\mathcal{M}},\Psi^{\mathcal{M}}) \log P(\mathbf{x}_i|\mathbf{y}_i,\theta^{\mathcal{M}})d\mathbf{y}_i\Big\}.\label{eq:maxtheta}
\end{array}\end{equation}
\begin{equation}\begin{array}{rl}
\beta^{\mathcal{M}}= \arg\max\Big\{\sum_{i=1}^T\int_{\mathbf{y}_i}P(\mathbf{y}_i|\mbf{x}_i,\mathbf{m}_i^{\mathcal{M}},\Psi^{\mathcal{M}})\log P(\mathbf{y}_i|\mathbf{m}_i^{\mathcal{M}},\beta^{\mathcal{M}})d\mathbf{y}_i\Big\}.\label{eq:maxbeta}
\end{array}\end{equation}
Subsequently, we maximise the log-likelihoods wrt. the parameters, recovering the update equations.  For $\theta$, by maximising Eq. \ref{eq:maxtheta}, we obtain
\begin{eqnarray}
\mathbf{W}^{\mathcal{M}}=\left( \sum_{i=1}^T \mathbf{x}_i\mathbb{E}^{\mathcal{M}}[\mathbf{y}_i]^T\right)\left(\sum_{i=1}^T\mathbb{E}^{\mathcal{M}}[\mathbf{y}_i\mathbf{y}_i^T]\right)^{-1}
\label{eq:update_W}
\end{eqnarray}
\begin{eqnarray}\begin{array}{rl}
	\sigma_x^{\mathcal{M}^2}&=\frac{1}{FT}\sum_{i=1}^T\{||\mathbf{x}_i||^2 - 2\mathbb{E}^{\mathcal{M}}[\mathbf{y}_i]^T(\mathbf{W}^{\mathcal{M}})^T\mathbf{x}_i\\
&+\mbox{Tr}[\mathbb{E}^{\mathcal{M}}[\mathbf{y}_i\mathbf{y}_i^T](\mathbf{W}^{\mathcal{M}})^T\mathbf{W}^{\mathcal{M}}]\}.
\end{array}
\label{eq:update_sigma_x}
\end{eqnarray}
Similarly, by maximising Eq. \ref{eq:maxbeta} for $\beta$, we obtain:
 \begin{equation}
\sigma_n^{\mathcal{M}^2} =\frac{F^{\mathcal{M}}(\lambda_n)}{T}\sum_{i=1}^T(\mathbb{E}^{\mathcal{M}}[y_{n,i}^2]
-2\mathbb{E}^{\mathcal{M}}[y_{n,i}]m_{n,i}^{\mathcal{M}} + m_{n,i}^{\mathcal{M}^2})
\label{eq:update_sigma_n}
\end{equation}
\looseness-1where, as defined in Eq. \ref{eq:genMultipleMRFpriorVar}, for PCA $F^\mathcal{M}(\lambda_n)=1$, and for LDA and LPP $F^\mathcal{M}(\lambda_n)=\lambda_n + (1-\lambda_n)^2$.
\if\uniFirst1
\else  
where
\begin{equation}
\begin{array}{rl}
\zeta^{\mathcal{M}}& = 
 \left\{
\begin{array}{l}
1 \;\mbox{for}\; R=\textrm{PCA} \\
\lambda_n+(1-\lambda_n)^2 \;\mbox{for}\; R=\textrm{LDA},\;R=\textrm{LPP}
\end{array}
\right.
\\
\end{array}\label{eq:meanfield}
\end{equation}
\fi
For $\lambda_n$ we choose the updates as described in Sec. \ref{sec:ML}.  In what follows, we discuss some further points wrt. the proposed EM framework.

\subsection{Further Discussion}
\label{sec:EM_discussion}

\looseness-1{\hspace{10pt} 
{\bf Comparison to other probabilistic variants of PCA.} 
It is clear that regarding the proposed EM-PCA, the updates for $\theta=\{\mbf{W}, \sigma_x^2\}$ as well as the distribution of the latent variable $\mbf{y}_i$ are the same with previously proposed probabilistic approaches \cite{roweis1998algorithms},\cite{TippingPPCA}.  The only variation is the mean of $\mbf{y}_i$, which in our case is shifted by the mean field, $\hat{\mathbf{{\Sigma}}}^{(\textrm{PCA})^{-1}}\mathbf{} \mathbf{m}_i^{\textrm{(PCA)}}$, while in addition, our method models per-dimension variance ($\sigma_n$).  Note that in order to fully identify with the PPCA proposed in \cite{TippingPPCA}, we can set $\lambda_n=0$ and $\sigma_n=1$.   

\looseness-1{\bf EM for SFA.}
\looseness-1The SFA prior in Eq. \ref{eq:prior:SFA} allows for two interpretations of the SFA graphical model: both as an undirected MRF and a directed Dynamic Bayesian Network (DBN).  Based on the undirected MRF interpretation, SFA trivially fits into the EM framework described in this section,  
leading to an auto-regressive SFA model \cite{rue2004gaussian}, able to learn bi-directional latent dependencies.
When considering the SFA prior as a directed Markov chain, one can resort to exact inference techniques applied on DBNs.  In fact, the EM for SFA can be straightforwardly  reduced to solving a standard Linear Dynamical System (Chap. 13 \cite{BishopBook}), while also enforcing diagonal transition matrices and setting $\sigma_n^2=1-\lambda_n^2$.

\begin{figure*}
\centering
\includegraphics[scale=0.75]{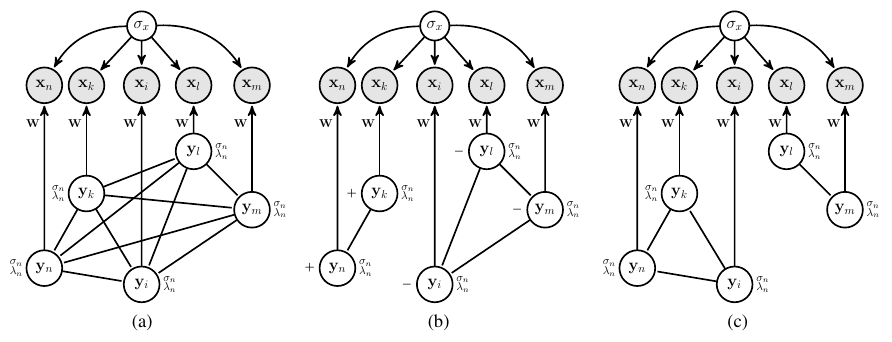}
\caption{MRF connectivities used for PCA, LDA and LPP under our unifying framework, with shaded nodes representing observations.  (a) Fully connected MRF (PCA), (b) within-class connected MRF (LDA), and (c) a locally connected MRF (LPP).\label{fig:motivating}}

\end{figure*}

{\bf Complexity.}
\looseness-1The proposed EM algorithm iteratively recovers the latent space preserving the characteristics enforced by the selected latent neighbourhood.  Similarly to PCCA \cite{roweis1998algorithms,TippingPPCA},  
for $N \ll T,F$ the complexity of each iteration is bounded by $O(TNF)$, unlike deterministic models ($\mathcal{O}(T^3)$).  This is due to the covariance appearing only in trace operations, and is of high value for our proposed models, especially in case where no other probabilistic equivalent exists.

{\bf Probabilistic LDA Classification.}
We can exploit the probabilistic nature of the proposed EM-LDA in order to probabilistically infer the most likely class assignment for unseen data.  Instead of using the inferred projection, we can essentially utilise the log-likelihood of the model.  In more detail, we can estimate the marginal log-likelihood for each test point $\mbf{x^*}$ being assigned to each class $c$: 
\begin{equation}\begin{array}{r}
\arg_{\,c}\max \left\{\log{P(\mbf{x}^*|\mbf{m}^{\mathcal{M}_c},\Psi^{\mathcal{M}})}\right\}
\end{array}\end{equation}
where by adopting the usual EM bound (as shown in Eq. \ref{eq:maxall}) this boils down to
\begin{equation}\begin{array}{r}
\arg_{\,c}\max \int_{\mathbf{y}^*_i} P(\mathbf{y}^*_i|\mbf{x}^*_i,\mathbf{m}^{\mathcal{M}_{c}},\Psi^{\mathcal{M}})\log P(\mathbf{x}^*_i,\mathbf{y}^*_i|\Psi^{\mathcal{M}})d\mathbf{y}^*_i
\end{array}\end{equation}
\looseness-1where $P(\mathbf{y}^*_i|\mbf{x}^*_i,\mathbf{m}^{\mathcal{M}},\Psi^{\mathcal{M}})$ is estimated as in Eq. \ref{eq:the_latent_posterior}, by utilising the inferred model parameters ($\Psi^{\mathcal{M}}$) along with the class model.  Note that since the posterior mean given $\mbf{x}_i$ depends on all {\it other} observations excluding $i$ (Eq. \ref{eq:EM:means_mi}), we only need to store the class mean estimated as a weighted average of all training data and all training data in class $c$, as
\begin{equation}
\mathbf{m}^{\mathcal{M}_{c}} = \mathbf{\Lambda}^{(\alpha)} \frac{1}{T}\sum_{j=1}^T \mathbb{E}^{\mathcal{M}}[\mbf{y}_j] + \mathbf{\Lambda}^{(\beta)} \frac{1}{|\mathcal{C}_c|}\sum_{j\in \mathcal{C}_c} \mathbb{E}^{\mathcal{M}}[\mbf{y}_j]
\end{equation}
This is in contrast to traditional methods where all the (projected) training data have to be kept.  Furthermore, during evaluation, we only need to estimate the likelihood of each test datum's assignment to each class ($\mathcal{O}(|{C}|$), rather than compare each test datum to the entire training set $(\mathcal{O}(T))$.

\section{Experiments}
\label{sec:experiments}
\looseness-1As proof of concept, we provide experiments both on synthetic and real-world data.  We aim to (i) experimentally validate the equivalence of the proposed probabilistic models to other models belonging in the same class, and (ii) experimentally evaluate the performance of our models against others in the same class.

{\bf Synthetic Data.} \looseness-1We demonstrate the application of our proposed probabilistic CA techniques on a set of synthetic data (see Fig. \ref{fig:synthetic}), generated utilising the Dimensionality Reduction Toolbox.  In more detail, we compare the corresponding deterministic formulations of PCA, LDA and LLE to the proposed probabilistic models.  The aim is mainly to qualitatively illustrate the equivalence of the proposed methods (by observing that the probabilistic projections match the deterministic equivalents).  Furthermore, the variance modelling per latent dimension in our EM-LDA is clear in $\mathbb{E}[\mbf{y}]$ of LDA (Fig. \ref{fig:synthetic}, Col. 3). This will prove beneficial prediction-wise, as we show in the following section. 
\begin{figure*}
\centering
\includegraphics[scale=0.73]{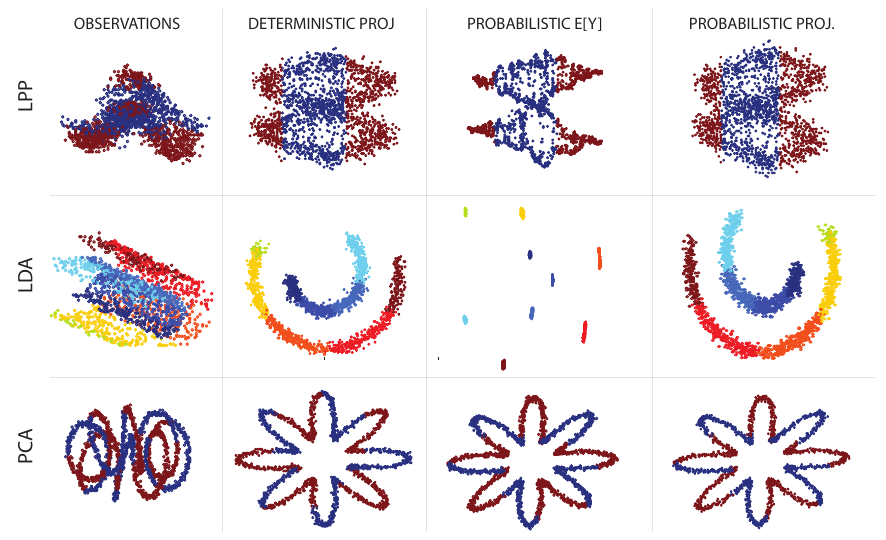}
\caption{\looseness-1 Synthetic experiments on deterministic LLE, LDA and PCA (2nd col.) compared to the proposed probabilistic methods ($\mathbb{E}[y]$ in 3rd col., projections in 4th col.).\label{fig:synthetic}}
\end{figure*}

{\bf Real Data: Face Recognition via EM-LDA. }One of the most common applications of LDA is face recognition.  Therefore, we utilise various databases in order to verify the performance of our proposed EM-LDA.  In more detail, we utilise the popular Extended Yale B database \cite{GeBeKr01}, as well as the PIE \cite{Sim_2002_3907} and AR databases \cite{martinez1998ar}.  The experiments span a wide range of variability, such as various facial expressions, illumination changes, as well as pose changes. 
In more detail from the CMU PIE database \cite{Sim_2002_3907} we used a total of 170 images near frontal images for each subject.  For training, we randomly selected a subset consisting of 5 images per subject, while for testing the remaining images were used.  For the extended Yale B database \cite{GeBeKr01}, we utilised a subset of 64 near frontal images per subject, where a random selection of 5 images per subject was used for training, while the rest of the images where used for testing. Regarding AR \cite{martinez1998ar}, we focus on facial expressions.  We firstly randomly select 100 subjects.  Subsequently, use the images which portray varying facial expressions from session 1, while using the corresponding images from session 2 for testing.
In related experiments, we compared our EM-LDA against deterministic LDA, the Fukunaga-Koontz variant (FK-LDA) \cite{zhang2007discriminant} and PLDA \cite{PLDAidentity} (which has been shown to outperform other probabilistic methods such as \cite{IoffeECCV2006} in \cite{li2012probabilistic})  under the presence of Gaussian noise.   We used the gradients of each image pixel as features, since as we experimentally verified, this improved the results for all compared methods.  The errors of each compared method for each database, accompanied by increasing Gaussian noise in the input, is shown in Fig. \ref{fig:face_rec_lda}.  Although PLDA offers a substantial improvement wrt. deterministic LDA and performs better than FK-LDA, it is clear that the proposed EM-LDA outperforms other compared LDA variants. This can be attributed to the explicit variance modelling (both for observations and per dimension) in our models, which appears to enable more robust classification.

\begin{figure*}[h]
\begin{center}
\hspace*{-10pt}\includegraphics[scale=1]{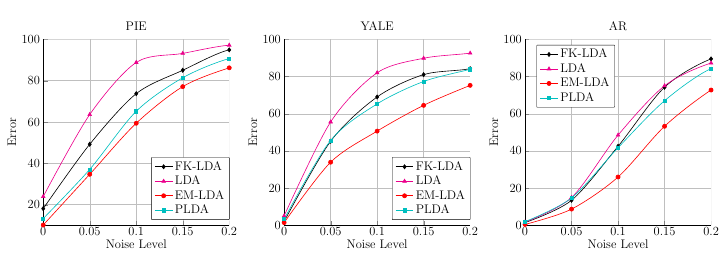}
\end{center}
\caption{Recognition error on PIE, YALE and AR under increasing Gaussian noise, comparing LDA, FK-LDA \cite{zhang2007discriminant} the proposed EM-LDA and PLDA \cite{PLDAidentity}.\label{fig:face_rec_lda}}
\end{figure*}
{\bf Real Data: Face Visualisation via EM-LPP.} One of the typical applications of Neighbour Embedding methods is the visualisation of, usually high-dimensional, data at hand. In particular, LPPs have often been used in visualising faces, providing an intuitive understanding of the variance and structural properties of the data \cite{roweis1998algorithms}, \cite{he2005face}.  In order to evaluate the proposed EM-LPP, which is to the best of our knowledge the first probabilistic equivalent to LPP \cite{niyogi2004locality}, we experiment on the Frey Faces database \cite{roweis2000nonlinear}, which contains 1965 images, captured as sequential frames of a video sequence.  We apply a similar experiment to \cite{he2005face}.  We firstly perturbed the images with random Gaussian noise, while subsequently we apply EM-LPP and LPP.  The resulting space is illustrated in Fig. \ref{fig:LPP}.  It is clear that the deterministic LPP was unable to cope with the added Gaussian noise, failing to capture a meaningful data clustering.  Note that the proposed EM-LPP was able to well capture the structure of the input data, modelling both pose and expression within the inferred latent space.
\begin{figure*}
\centering
\includegraphics[scale=0.63]{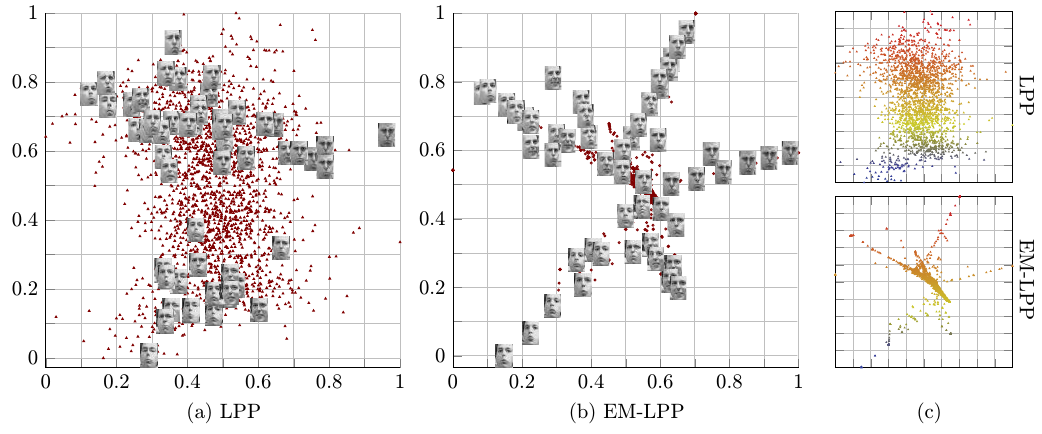}
\caption{Latent projections obtained by applying the proposed EM-LPP and LPP \cite{niyogi2004locality} to the Frey Faces database, with each image perturbed with random Gaussian noise.\label{fig:LPP}}
\end{figure*}
\section{Conclusions}
\looseness-1In this paper we introduced a novel, unifying probabilistic component analysis framework, reducing the construction of probabilistic component analysis models to selecting the proper latent neighbourhood via the design of the latent connectivity.  Our framework can thus be used to introduce novel probabilistic component analysis techniques by formulating new latent priors as products of MRFs.  We have shown specific priors which when used, generate probabilistic models corresponding to PCA, LPP, LDA and SFA, and by doing so, we introduced the first, favourable complexity-wise, probabilistic equivalent to LPP.  Finally, by means of theoretical analysis and experiments, we have demonstrated various advantages that our proposed methods pose against existing probabilistic and deterministic techniques.
{
\footnotesize
\section{Acknowledgements}
This work has been funded by the European Union's 7th Framework Programme [FP7/2007-2013] under grant agreement no. 288235 (FROG), the EPSRC project EP/J017787/1 (4DFAB) and the European Community 7th Framework Programme [FP7/2007-2013] under grant
agreement no. 611153 (TERESA).
}

{
\footnotesize
\bibliographystyle{splncs03}
\bibliography{UniProbComp}
}
\end{document}